\documentclass[letterpaper, 10pt, conference]{ieeeconf}      
 
\IEEEoverridecommandlockouts                              

\usepackage[colorlinks,bookmarksopen,bookmarksnumbered]{hyperref}

\usepackage{color}
\usepackage{graphicx}
\usepackage{amsmath}
\usepackage{amssymb}
\usepackage{tabularx}
\usepackage{subcaption}
\usepackage{url}
\usepackage{cite}
\usepackage{prettyref}
\newcolumntype{P}[1]{>{\centering\arraybackslash}p{#1}}

\DeclareMathOperator*{\argmax}{\arg\!\max}

\DeclareMathOperator{\softmax}{softmax}

\newcommand{\E}[1]{\mathbf{#1}}

\newrefformat{fig}{Figure~\ref{#1}}
\newrefformat{sec}{Section~\ref{#1}}
\newrefformat{tab}{Table~\ref{#1}}
\newrefformat{alg}{Algorithm~\ref{#1}}
\newrefformat{eq}{Equation~\ref{#1}}

\title{\LARGE \bf 
Safe Navigation with Human Instructions in Complex Scenes}
\author{Zhe Hu, Jia Pan$^\dagger$, Tingxiang Fan, Ruigang Yang and Dinesh Manocha%
\thanks{Z. Hu and J. Pan are with the Department of Mechanical and Biomedical Engineering, the City University of Hong Kong, Hong Kong. T.X. Fan and R. Yang are with the Robotics and Autonomous Driving Lab, Baidu Research. D. Manocha is with the Department of Computer Science, the University of Maryland, College Park. $^\dagger$ denotes the corresponding author. Email: jiapan@cityu.edu.hk}%
}

\begin{document}
\maketitle

\begin{abstract}
In this paper, we present a robotic navigation algorithm with natural language interfaces, which enables a robot to safely walk through a changing environment with moving persons by following human instructions such as ``go to the restaurant and keep away from people''. We first classify human instructions into three types: the goal, the constraints, and uninformative phrases. Next, we provide grounding for the extracted goal and constraint items in a dynamic manner along with the navigation process, to deal with the target objects that are too far away for sensor observation and the appearance of moving obstacles like humans. In particular, for a goal phrase (e.g., ``go to the restaurant''), we ground it to a location in a predefined semantic map and treat it as a goal for a global motion planner, which plans a collision-free path in the workspace for the robot to follow.
For a constraint phrase (e.g., ``keep away from people''), we dynamically add the corresponding constraint into a local planner by adjusting the values of a local costmap according to the results returned by the object detection module. The updated costmap is then used to compute a local collision avoidance control for the safe navigation of the robot.  
By combining natural language processing, motion planning and computer vision, our developed system is demonstrated to be able to successfully follow natural language navigation instructions to achieve navigation tasks in both simulated and real-world scenarios. Videos are available at \url{https://sites.google.com/view/snhi}.
\end{abstract}


\section{Introduction}
In the human-robot interaction (HRI) field, natural language has become an important communication interface between humans and robots~\cite{wei2009go,kollar2014grounding,kollar2010toward,posada2014visual}. 
Many previous works endow robots with the ability to understand human instructions by using probabilistic models to ground a human instruction to a robot action~\cite{kollar2013generalized,howard2014natural,arkin2015towards,chung2015performance,paul2016efficient}.
However, even though demonstrated to be successful in manipulation tasks, these methods may not be appropriate for navigation tasks. This is because compared to a manipulation task, a navigation task would have a much larger configuration space, which will result in the training and inference difficulty of these direct grounding approaches. To deal with such difficulty, one feasible solution is using the motion planning as an intermediate layer bridging the natural language processing (NLP) module and the action generation module. Similar ideas have been proposed before in~\cite{park2017generating} for manipulation tasks. In particular, we will use NLP techniques to extract semantic information such as goals and constraints of the navigation task from human instructions. Such knowledge is then fed into a costmap motion planner to compute an optimal navigation trajectory. In this way, we can efficiently solve the action grounding problem for navigation tasks.

\begin{figure}[t]
\centering
\includegraphics[trim=20 60 30 0, clip, width=\linewidth]{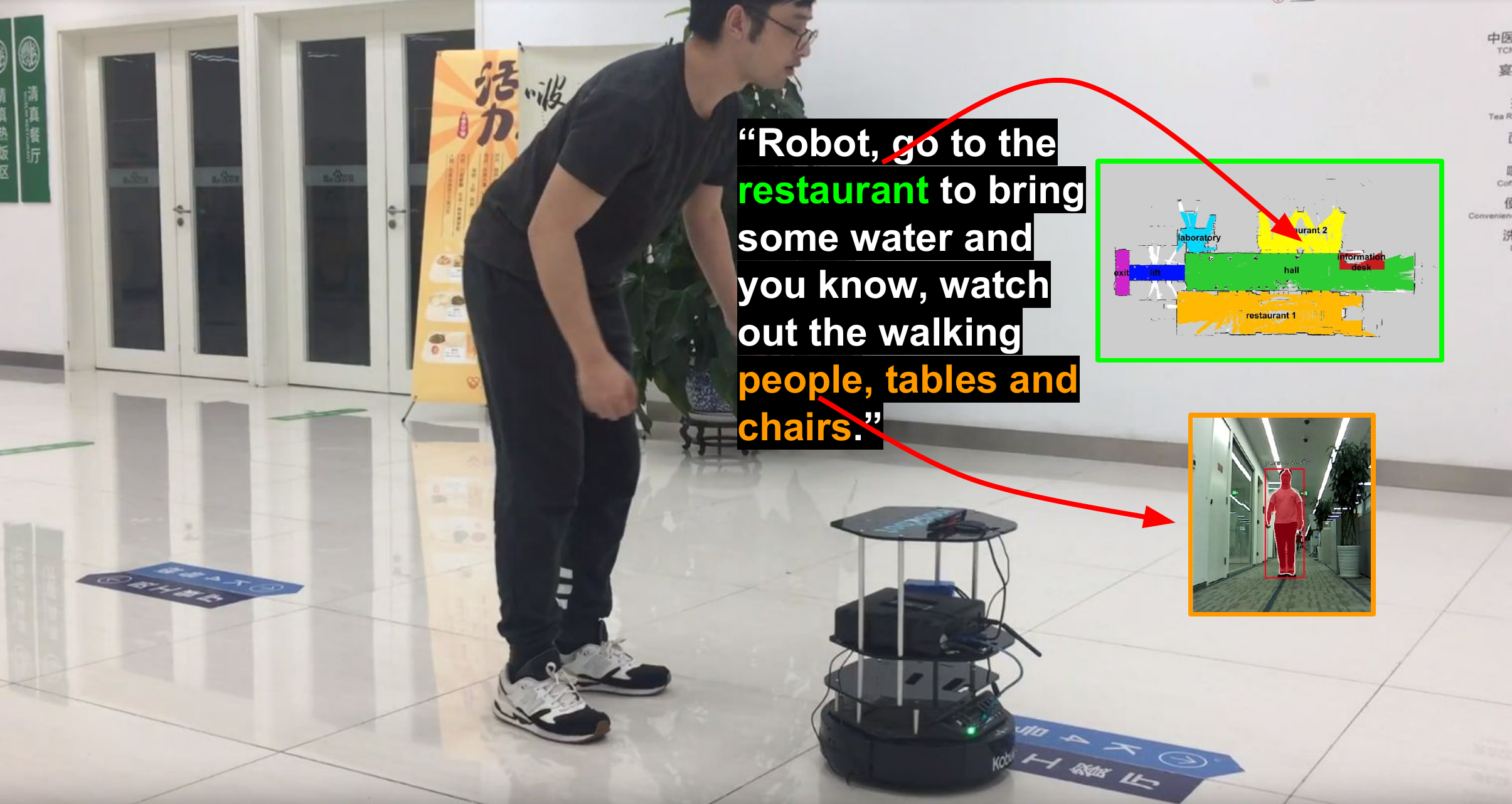}
\caption{\label{fig:scenario} Our system enables a robot to understand the goals and constraints in the natural language instruction and localize these abstract concepts with concrete objects in the physical world by using dynamic grounding techniques, which leverage the knowledge from a global semantic map (in the green box) and from the RGB-D sensor (in the orange box). The grounded knowledge is then fed into a costmap motion planner to accomplish global navigation in complex scenarios with both static and moving obstacles. }
\vspace*{-0.12in}
\end{figure}

However, the semantic output of the NLP module cannot be directly used by the motion planner, which requires concrete knowledge about positions of the goal and the objects referred to by constraints. For instance, given the constraint phrase ``keep far away from the red desk'' that is extracted from the NLP module, the motion planning algorithm needs to understand what is the item ``red desk'' referred to in the physical world, and what is its position in the robot's local coordinate system. In other words, we need to ground abstract semantic concepts from the NLP module to concrete physical attributes of the objects that are ready to be used in the motion planning framework. Motivated by the visual grounding technique~\cite{huang2018finding,shridhar2018interactive}, which takes image-sentence pairs as input and learns to visually localize linguistic phrases with pixels or regions in a given image, our solution is to extract the concrete positions of goals and constraint-related objects from RGB-D images, by combining computer vision and NLP techniques. In addition, due to occlusion and the camera's limited range, the robot usually cannot ground all constraints at a time. As a result, the robot needs to dynamically analyze the surrounding environment for possible grounding and add a constraint into the motion planner once it has been grounded.

\noindent{\bf Main Results:}
In this paper, we present an algorithm to enable a mobile robot to understand the navigation goal and trajectory constraints from natural language human instructions and sensor measurements, in order to generate a high-quality navigation trajectory following user's requirements. 
We first use a Long Short-Term Memory (LSTM) recursive neural network to parse and interpret the commands and generate the navigation goal and a set of constraint phrases. The navigation goal is in form of a location name in the semantic map. The constraint phrase describes the spatial relationship between the robot and a target object, e.g., in ``keep away from the desk'', ``desk'' is the target object and ``keep away from'' is the spatial relation. Moreover, we use the attention mechanism to improve the accuracy of the NLP command parsing. Next, we perform a dynamic grounding algorithm to localize the extracted linguistic phrases with concrete items in the physical world. In particular, we will link the goal phrase to the corresponding physical location in the semantic map and will link the target object for a constraint phrase to a point cloud observed by the RGB-D sensor. In this way, we can translate the abstract NLP commands into parameters that can be used in a motion planner. For instance, ``keep away from the desk'' can be translated into a costmap $f(\mathbf x) = 1_{\|\mathbf x - \mathbf x_{\text{desk}}\|\leq r}(\mathbf x)$, where $\mathbf x$ is one point in the local costmap, $\mathbf x_{\text{desk}}$ is the grounded position for the desk, $1(\mathbf x)$ is the indicator function, and $r$ is the potential field parameter. The constraints will be grounded whenever the target object is observed by the robot, and thus the costmap is dynamically updated during the navigation. 
Finally, a trajectory is computed using an online search over a costmap in the robot's local coordinate system. The planned trajectory is further combined with a local collision avoidance module to enable the robot approach its goal safely and efficiently.
As compared to prior techniques, our approach offers the following benefits:
\begin{itemize}
\item We use the costmap motion planner as the bridge connecting the NLP module and the navigation command generator. The costmap can conveniently formulate parameters like goal positions and spatial constraints, and also allows the dynamic update of these parameters. 
\item Our dynamic grounding allows constraint phrases containing moving objects, e.g., ``keeping away from people''. By combining with our collision avoidance technique, it can provide safe navigation in crowds. 
\item We use LSTM enhanced with attention mechanisms to provide high-quality and efficient NLP information extraction for the navigation task.
\end{itemize}
We highlight the performance of our method in a simulated environment as well as on a Turtlebot operating in two real-world scenarios. Our approach can handle a rich set of natural language commands and can generate appropriate paths in real-time.

\section{Related Work}
\label{sec:related}

\begin{figure*}[!htb] 
\centering
\includegraphics[trim=1 120 5 110, clip, width=\linewidth]{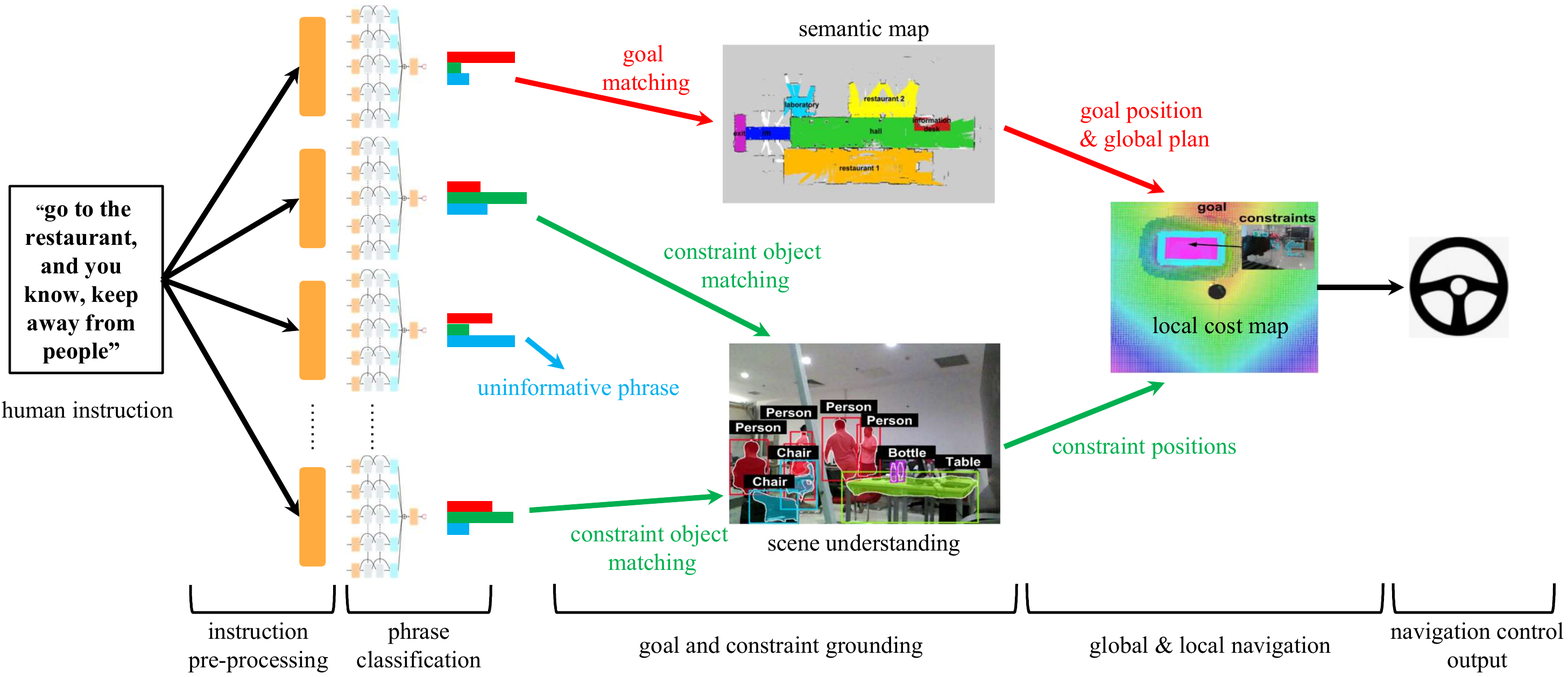}
\caption{An overview of our navigation algorithm, which takes the human instructions as input and generates a suitable trajectory. The algorithm has four main modules: the instruction pre-processing, the phrase classification, the goal and constraint grounding, and the global and local navigation planning. The instruction pre-processing splits the speaker input into a set of phrases. The phrase classification uses a bi-directional LSTM enhanced using attention mechanisms to understand the function of each phrase. Given a phrase, the LSTM output is a class probability about whether this phrase is a goal indicator (the red bar), a constraint descriptor (the green bar), or uninformative words (the blue bar). The goal and constraint grounding module maps the goal and constraint phrases to concrete objects in the physical world. In particular, we determine the goal's location by search the goal name in the semantic map, and determine the target object in a constraint by matching with the output from the scene understanding. The grounded goal and constraints are then used to update the cost map for computing a navigation control output.}
\label{fig:overview}
\vspace*{-0.2in}
\end{figure*}

\subsection{Natural Language Processing in Robotic Navigation}
Many previous works introduce natural language into robotic navigation and use probabilistic graph models as a key tool to perform language grounding. Wei et al.~\cite{wei2009go} presented a variant of Markov Random Field (MRF) to perform path inference from human instructions. Kollar et al.~\cite{kollar2014grounding}~\cite{kollar2010toward} proposed an algorithm called Spatial Description Clauses (SDCs) to transform instructions into a structured formulation and then used a probabilistic model to perform verb and direction grounding. Tellex et al.~\cite{tellex2011understanding} also used SDCs to describe instructions but presented some useful new features, such as distance, to formulate the probabilistic model. Kollar et al.~\cite{kollar2013generalized} presented a special probabilistic framework called the Generalized Ground Graphs ($\mathbf{G}^3$) to ground a word phrase to an object, place, path, or event in the physical world, where the probabilistic graph is a factor graph constructed based on SDCs. Once the graph is built and trained, the grounding can be obtained by maximizing the conditional probability described by the graph. Howard et al.~\cite{howard2014natural} extended the $\mathbf{G}^3$ model to the Distributed Correspondence Graph (DCG) by adding all possible grounding nodes to reduce the search space. Instead of grounding actions, this work directly grounds constraints to obtain better inference performance. Built upon a similar idea, our approach also infers the constraints directly and combines motion planning with NLP computations.~\cite{arkin2015towards,chung2015performance} proposed a variant of DCG called the Hierarchical Distributed Correspondence Graph (HDCG) for dealing with the computation of large and complex symbolic representations. Paul et al.~\cite{paul2016efficient} extended DCG and proposed a model called the Adaptive Distributed Correspondence Graph (ADCG), which provides a hierarchy of the meanings attached to a word phrase to deal with abstract groundings. Park et al.~\cite{park2017generating} presented a similar probabilistic graph called the Dynamic Grounding Graphs (DGG) to dynamically parse and interpret commands and generate constraints for optimization-based motion planning. Their work modeled the motion planner's parameters as latent variables and used Conditional Random Fields (CRF) to perform training and inference. There are also some other works using techniques other than the probabilistic graphs. Matuszek et al.~\cite{matuszek2010following} used a statistical machine translation model that transforms natural language instructions to an automatically-labeled map, which is then used for navigation. Duvallet et al.~\cite{duvallet2013imitation} applied imitation learning to the direction following task, which used demonstrations to learn a policy about how to explore the environment, how to backtrack when necessary, and how to explicitly declare when the
destination has been reached. The learned policy can be well generalized to unknown environments. 

\subsection{Visual Grounding}
There is a rich literature about visual grounding, i.e., how to visually ground (i.e., localize) arbitrary linguistic phrases with pixels or objects in an image. Huang et al.~\cite{huang2018finding} built a connection graph between a word phrase and a visual bounding box and then performed inference using EM (Expectation Maximization) to ground referring expressions like ``it''. Shridhar et al.~\cite{shridhar2018interactive} proposed to solve the referring expression grounding problem by combining Fast R-CNN with LSTM to generate a word probability sequence, which represents an expression distribution describing the image region. Anderson et al.~\cite{anderson2018vision} proposed a simulation environment for visually-grounded navigation and presented an LSTM based reinforcement learning algorithm to perform simulated room-to-room navigation. Their LSTM formulation not only takes previous images as input but also considers actions taken in the previous steps. This mechanism allows their model to handle more complex instructions than previous approaches. Wu et al.~\cite{wu2018building} proposed a 3D navigation environment called House3D, which includes 3D objects, textures, and scene layouts. Their dataset can generate both low-level and high-level variations like color and layout changes. Matuszek et al.~\cite{matuszek2012joint} proposed a joint probabilistic model of language and perception for visual grounding. Posada et al.~\cite{posada2014visual} presented a robotic navigation framework without a global map. Their approach performs object segmentation and detection given scene images using a set of traditional classification modules such as Random Forest and Support Vector Machine.

\section{Overview}


The goal of our work is to enable robots to follow human instructions such as ``go to the restaurant, and you know, keep away from people.'' Our navigation system takes the command sentence as the input and outputs a suitable trajectory for robots to execute. Our approach contains four steps: pre-processing, phrase classification, goal and constraint grounding, and motion planning. First, the pre-processing step takes the command sentence as the input and divides it into phrases according to conjunctions and commas. Next, we assume that each phrase has one of three labels: goal, constraint, or uninformative phrase. For example, the phrase ``go to the restaurant'' is a goal phrase, ``keep away from people'' is a constraint phrase, and the phrase ``you know'' is uninformative. We train a Long Short-Term Memory (LSTM) network~\cite{hochreiter1997long} to classify phrases into those three types. The LSTM is suitable for our task because the classification output does not depend only on individual words but also on the meaning of the entire phrase.
After recognizing the constraint and goal phrases, we need to ground them with the physical world. In particular, we ground the goal phrase by computing the similarity between the noun extracted from the goal phrase and the location name in the predefined semantic map, and use the most similar location as the goal configuration for the robot navigation. To dynamically add the target object in a constraint into the planner's costmap, we use an object detection module to dynamically look for the object mentioned in the constraint phrase and then output the object's 3D location in the robot's local coordinate system. To deal with objects that are moving (like ``people'') or objects that are occluded, we will perform the grounding in a dynamic manner along with the navigation. That is, whenever a constraint object occurs in the camera frame, we add it to the local planner's costmap.
Finally, our costmap motion planner will compute a suitable trajectory based on the costmap which is updated in real-time according to the grounded constraint result. To improve the navigation performance in crowd scenarios, we further use the planning result to guide a reinforcement learning based local collision avoidance approach developed in our previous work~\cite{long2017towards}. An overview of our proposed navigation system is shown in \prettyref{fig:overview}.

\section{Phrase Classification}
\label{sec:phraseclassification}
To understand the roles that different instruction phrases play in the motion planning, we propose an LSTM-based method to ground phrases into different types. For navigation task, we notice that the goal and the user-specified trajectory constraints are the most important parameters for the planning algorithm. As a result, we classify each phrase into one of three types: goal, constraint, and uninformative phrase. For example, ``go to the restaurant'' is a goal phrase, ``keep away from people'' is a constraint phrase, and ``you know'' is an uninformative phrase. We train a bi-directional LSTM network~\cite{graves2005framewise} with attention mechanism to solve such classification problems. In \prettyref{sec:experiment}, we show that the bi-directional LSTM network with attention mechanism can provide better classification performance than standard LSTM.

\begin{figure}[!htb] 
\centering
\includegraphics[trim=7 65 8 65, clip,  width=0.9\linewidth]{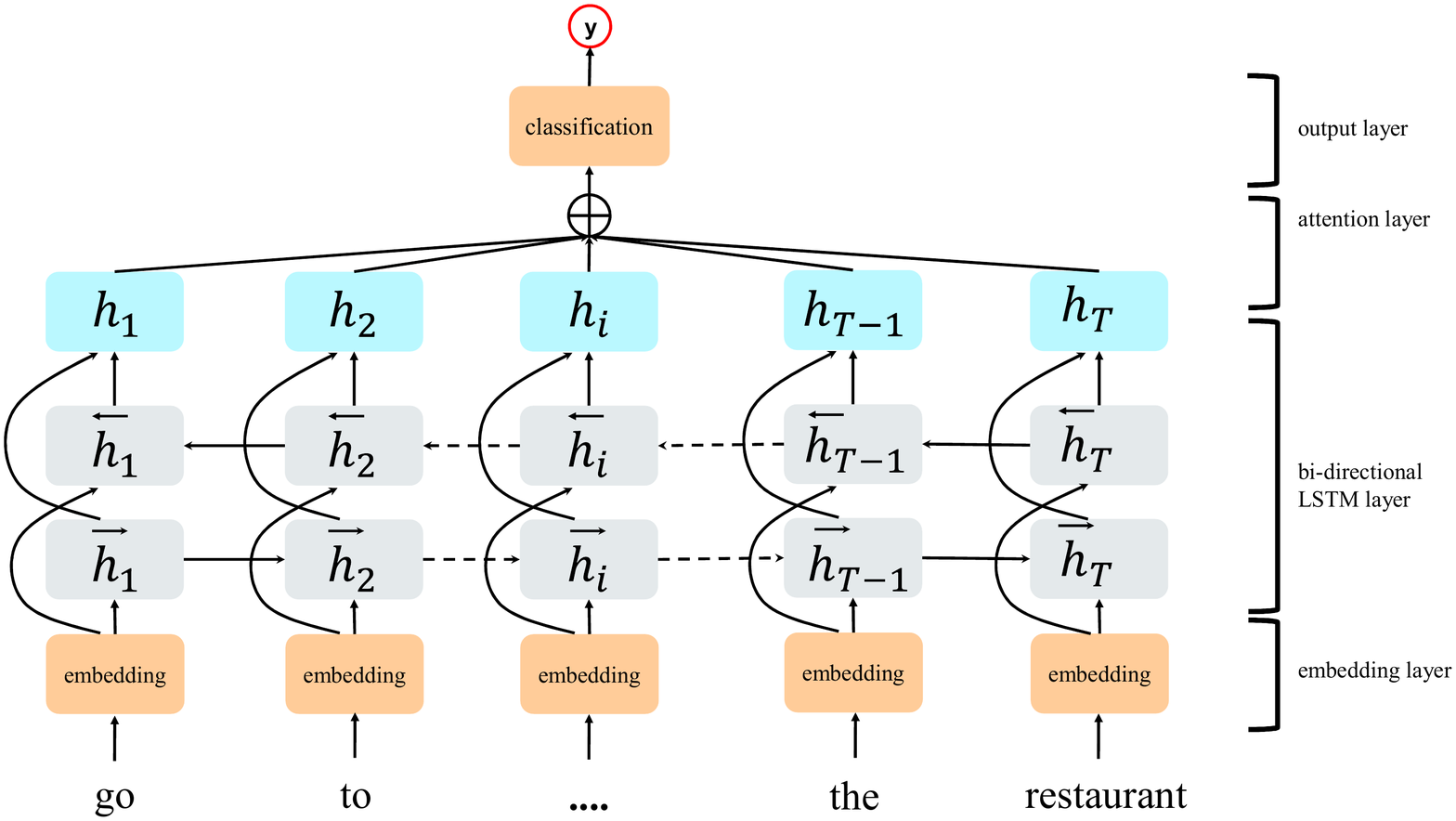}
\caption{The phrase classification bi-directional LSTM network.}
\label{fig:pc}
\centering
\includegraphics[trim=0 50 0 40, clip, width=0.4\linewidth]{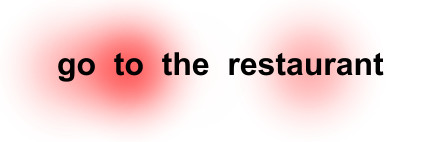}
\caption{Attention visualization for a goal phrase, where the color with higher transparency indicates a lower attention value.}
\label{fig:attention}
\end{figure}

More specifically, as shown in \prettyref{fig:pc}, our classification subsystem contains four layers: the embedding layer, the bi-directional LSTM layer, the attention layer, and the output layer. The embedding layer transforms a phrase $S$ with length $T$: $S = \{x_1, x_2, \cdots, x_T\}$ into a list of real-valued vectors $E = \{\E{e}_1, \E{e}_2, \cdots, \E{e}_T\}$. We transform a word $x_i$ into the embedding vector through an embedding matrix $\E{W}_e$: $\E{e}_i = \E{W}_e \E{v}^i$, where $\E{v}^i$ is a vector of value 1 at index $e_i$ and $0$ otherwise. 

Next, we extract the representation for the sequential data $E$ using LSTM, which provides an elegant way of modeling sequential data. But in standard LSTM, the information encoded in the inputs can only flow in one direction and the future hidden unit cannot affect the current unit. To overcome this drawback, we employ a bi-directional LSTM layer~\cite{graves2005framewise}, which can be trained using all the available inputs from two directions to improve the prediction performance. 

A bi-directional LSTM consists of a forward and backward LSTM. The forward LSTM reads the input embedded vector sequence from $e_1$ to $e_T$ and computes a sequence of forward hidden states $(\overrightarrow{\E{h}_1}, ..., \overrightarrow{\E{h}_T})$, where $ \overrightarrow{\E{h}_i} \in \mathbb{R}^p$ and $p$ is the dimensionality of hidden states. The backward LSTM reads the embedded vector sequence in the reverse order, i.e., from $\E{e}_T$ to $\E{e}_1$, resulting in a sequence of backward hidden states $(\overrightarrow{\E{h}_T}, ..., \overrightarrow{\E{h}_1})$, where $\overrightarrow{\E{h}_T}\in \mathbb R^p$. By adding the forward and backward states $\overrightarrow{\E{h}_i}$ and $\overleftarrow{\E{h}_i}$, we can obtain the final latent vector representation as 
$\E{h}_i = \overrightarrow{\E{h}_i} + \overleftarrow{\E{h}_i}$,
where $\E{h}_i \in \mathbb R^p$.

Given the hidden states of the bi-directional LSTM, we want to combine them to predict the class label of each phrase. Here we use the attention mechanism, which has been successfully applied in many tasks such as image captioning, language translation, speech recognition. Its success is mainly due to the fact that human recognition does not tend to process a whole signal at once; instead one only focuses on selective parts of the entire perception space as needed. For phrase classification, we add an attention layer after the bi-directional LSTM layer to weight hidden units differently so as to improve the classification performance. Here we are using a location-based attention mechanism, which calculates the weights solely from the current hidden state $\E{h}_i$ as $\alpha_i = \E{w}^T \tanh(\E{h}_i)$, where $\E{w} \in \mathcal R^p$ is a vector of parameters to be learned. We then can obtain an attention weight vector $\alpha$ using softmax function as follows:
\begin{equation}
\label{eq:attent_weight}
\alpha = \softmax([\alpha_1, ..., \alpha_T]).
\end{equation}
Then a summarized vector $\E{r} \in \mathcal R^p$ can be computed based on the weights obtained in \prettyref{eq:attent_weight} and the hidden states from $\E{h}_1$ to $\E{h}_T$ as $\E{r} = \sum_{i=1}^T \alpha_i \E{h}_i$.
The summarized vector then passes through a $\tanh$ nonlinear function to generate the final output $\E{h}^*$ of the attention layer:
$\E{h}^* = \tanh(\E{r})$.

In this way, different words can have different impacts on the final classification results. In \prettyref{fig:attention}, we illustrate the visualization of the learned attention for the goal phrase. As we can observe, the word ``to'' has a greater impact on the goal phrase and the word ``go'' is also related to the goal. In \prettyref{sec:experiment}, we further show that the attention mechanism can indeed improve the performance of our navigation algorithm.

The final layer in our phrase classification network is the output layer, which is a dense layer with 3 neural units and the softmax activation and outputs the probability $p(y)$ that the phrase belongs to each class:
\begin{equation}
p(y) = \softmax(\E{W} \E{h}^*+\E{b}),
\end{equation}
\begin{equation}
\hat{y} = \argmax_y p(y),
\end{equation}
where $\E{W}$ and $\E{b}$ are the parameters of the output neural layer to be learned, and $y$ is the phrase label from the set of goal, constraint, and uninformative phrase.
\section{Dynamic Goal and Constraint Grounding}
\label{sec:dcg}

For goal grounding, we compute the goal location by extracting the noun from the goal phrase and then computing the similarity between this noun and the location names in the predefined semantic map. The similarity is computed as the cosine similarity between the embedded vectors of two given word items. The embedding is accomplished using the Word2Vec embedding network~\cite{mikolov2013efficient,mikolov2013distributed} that can convert a word into a vector based on the Wiki corpus. We use the 2D coordinate of the location that has the highest similarity with the noun in the goal phrase as the goal location. 

For constraint grounding, we first determine whether a constraint object exists in the current scenario and if it does, determine its locations in the robot's local coordinate system. 
Previous works~\cite{wei2009go,paul2016efficient} used predefined objects and their locations to ground the language rather than dynamically determining their locations according to the output of a vision system. Our navigation algorithm combines NLP with computer vision to achieve dynamic constraint grounding, which can determine the existence and location of the constraint objects in an online manner.

\begin{figure}[!htb] 
\centering
\includegraphics[trim=0 0 0 145, clip, width=1\linewidth]{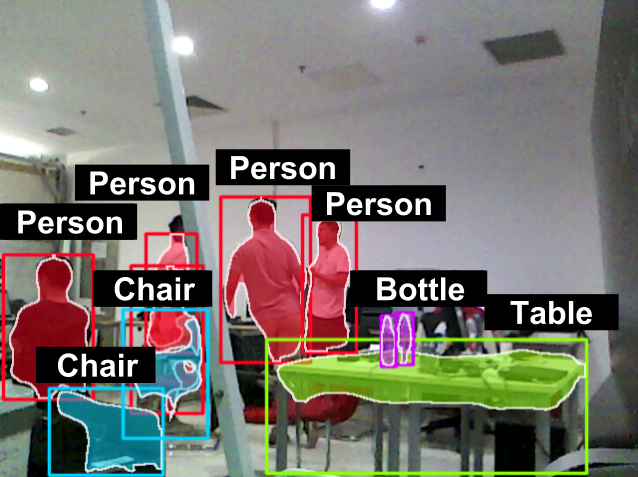}
\caption{For scene understanding, we use a mask R-CNN based instance segmentation to extract semantic objects in the scenario.}
\label{fig:seg}
\end{figure}

In particular, we first understand the current scene using instance segmentation, which takes one image frame as input and outputs several masks with semantic information (i.e., labels), as shown in \prettyref{fig:seg}. To compute the location of each object in the robot's local coordinate, we apply these masks to the corresponding depth image and compute the mean depth value over each mask, which is then used as the 3D location of each object.

Once we label the objects in the current scene, we compute the similarity between the nouns that occur in the constraint phrase and those object labels discovered by the scene understanding. Again, we use the Word2Vec embedding network~\cite{mikolov2013efficient,mikolov2013distributed} to embed all words in the vector space and compute the cosine similarity between word vectors of the constraint object and objects detected in the instance segmentation. If the similarity is larger than a predefined threshold, we add this constraint to the motion planner.

\section{Global and Local Planning}
\label{sec:pathplan}

For both global and local planning, we assume that the robot can accurately localize itself in a predefined global semantic map. We use a state-of-the-art 2D localization technique~\cite{cartographer} for the robot localization. The global semantic map is constructed using SLAM (simultaneous localization and mapping) algorithm. The resulting map consists of grid cells which may be one of three types: free space, obstacle, and no information.

\subsubsection{Global Planning in the Semantic Map}
To perform global planning, we use the RRT algorithm to compute a trajectory connecting the robot's current position in the semantic map and the goal position determined with goal grounding in \prettyref{sec:dcg}.

\subsubsection{Local Planning in the Costmap}
\begin{figure}[!htb] 
\centering
\begin{subfigure}{0.23\textwidth}
\includegraphics[width=1.0\linewidth]{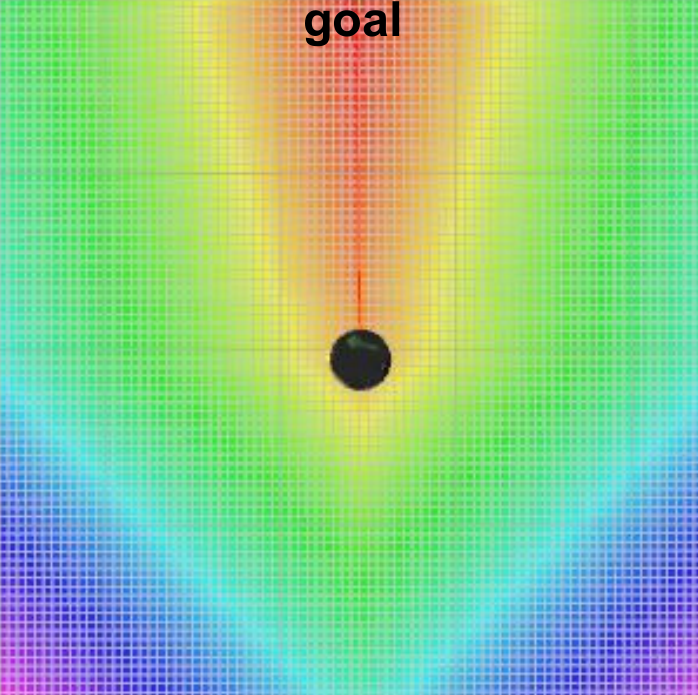}
\caption{}
\label{fig:constraint1}
\end{subfigure}
\begin{subfigure}{0.23\textwidth} 
\includegraphics[width=1.0\linewidth]{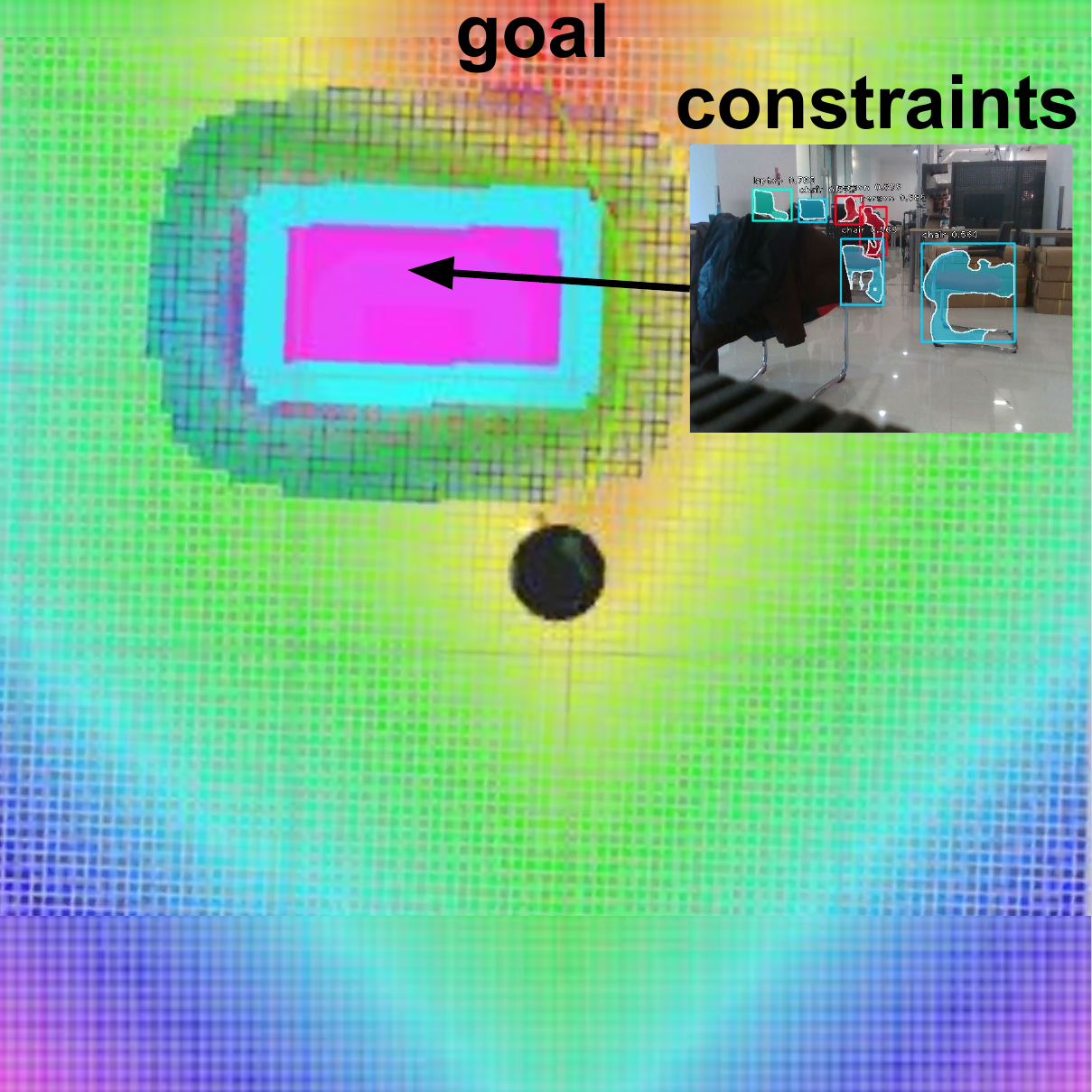}
\caption{}
\label{fig:constraint2}
\end{subfigure}
\caption{Path planning and the corresponding costmap visualization: (a) without constraints and (b) with constraints.}
\label{fig:mp}
\end{figure}

The robot then starts to follow the globally planned trajectory, but it needs to perform local planning to adapt to the dynamically added constraints. In particular, the local planning algorithm will maintain a costmap in the robot's local coordinate. The costmap contains both static obstacles and constraints. The static obstacles like walls, rooms, and buildings can be directly added into the costmap by transforming a subset of the global map. The constraints can also be conveniently modeled in the costmap. Given the location $(x_0,y_0)$ of a constraint object computed via constraint grounding, we can mark the cells inside the region $\{(x,y): (x-x_0)^2 + (y-y_0)^2 \leq a^2\}$ as the obstacle cells, where $a$ is a parameter indicating the influencing radius of the constraint. Finally, we perform the smooth inflation operation over all the obstacle cells in the grid map, in order to enable the robot to keep a safe distance from the obstacle. Samples of the costmap computed are shown in \prettyref{fig:mp}.

Given the costmap and the globally planned trajectory, the local planning is performed as follows. First, we choose one point from the global planned trajectory as the intermediate goal in the short time horizon. Next, we add this intermediate goal into the costmap by converting it to the local coordinate of the robot. After that, we use $A^*$ algorithm to compute an optimal path in the costmap, which is used as the local planning result. 

\subsubsection{Dynamic Obstacle Avoidance}
After global planning and local planning, the robot can avoid static obstacles (e.g., walls, heavy desks) that appear in the global semantic map and constraint obstacles that appear in the human instructions. However, there exist some other obstacles in the navigation environment, including static obstacles like chairs and non-static obstacles like carts and dogs, which appear neither in the human instruction nor in the semantic map. 
To guarantee the robot to avoid these obstacles reliably, our navigation algorithm combines the output of the local planning with a reinforcement learning-based local collision avoidance policy developed in our previous work~\cite{long2017towards}.

\begin{figure}
\centering 
\includegraphics[width=\linewidth]{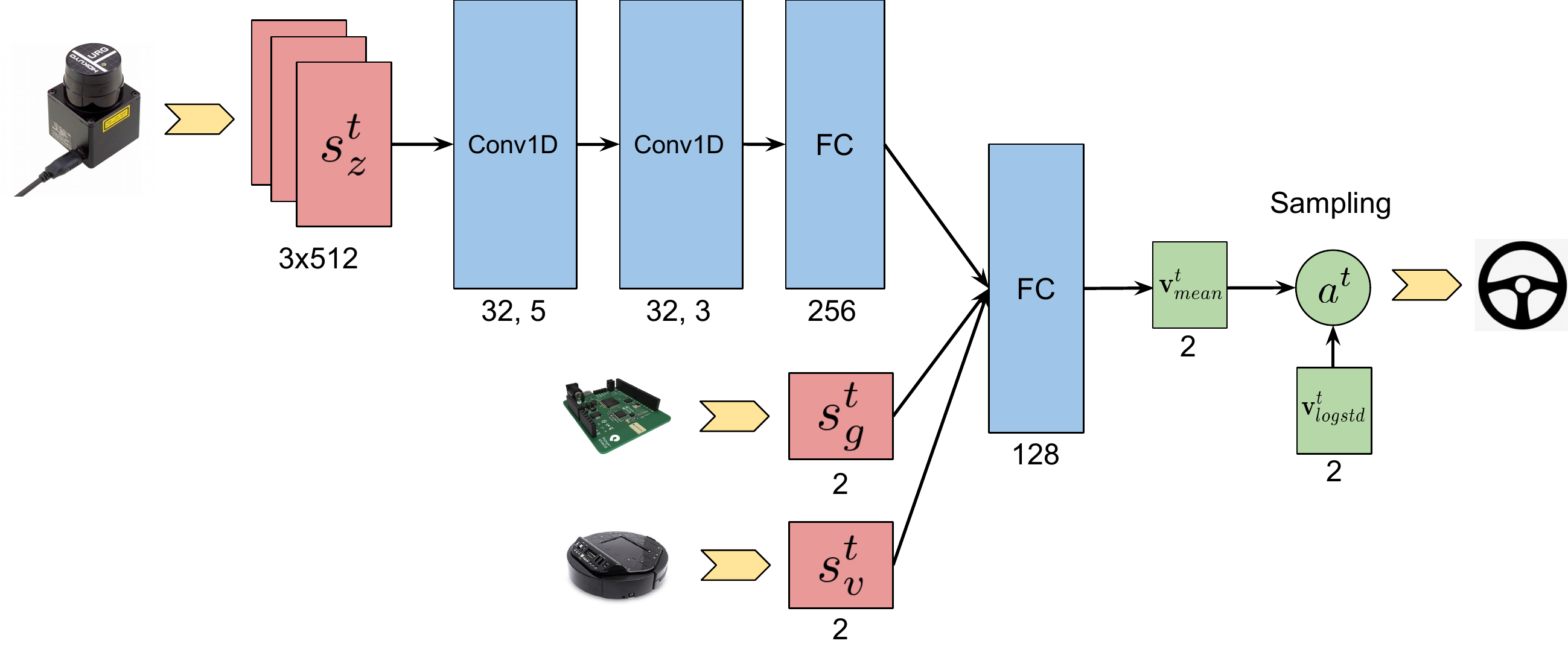} 
\caption{The neural network architecture. The network takes the laser scanner's data $\mathbf{o}^t_z$, relative goal position $\mathbf{o}^t_g$ and current velocity $\mathbf{o}^t_v$ as inputs, and outputs the mean of velocity $\mathbf{v}^t\textsubscript{mean}$. The executable action $\mathbf a^t$ is sampled from the Gaussian distribution constructed by $\mathbf{v}^t\textsubscript{mean}$ with a separated log standard deviation vector $\mathbf{v}^t\textsubscript{logstd}$.} 
\label{fig:model} 
\vspace*{-0.1in}
\end{figure}

The local collision avoidance controller is a 4-hidden-layer neural network as shown in \prettyref{fig:model}. It requires three inputs: the sensor scanning about the surrounding environment, the current velocity of the robot, and an intermediate goal for the robot to approach. We train the neural network policy using a wide variety of multi-agent scenarios, in order to enable the robot to learn a sophisticated coordination behavior so that accomplish safe and efficient navigation in scenarios with high agent density and complex static obstacles. The training is implemented as an extension of the state-of-the-art reinforcement learning algorithm PPO (Proximal Policy Optimization)~\cite{schulman2017proximal}. For more details about the collision avoidance controller, please refer to~\cite{long2017towards}.

In our navigation algorithm, the local collision avoidance controller will use the waypoints generated by the local planning as the goal input, and then outputs a navigation command which will drive the robot toward the eventual goal but also avoid all the obstacles in the environment.

\section{Experiments and Results}
\label{sec:experiment}
We test the performance of our natural language driven navigation algorithm on a Turtlebot robot platform mounted with an Intel Core i5-7500T CPU and a GeForce GTX 1070 GPU. We first present the experimental results on individual modules including phrase classification and dynamic constraint grounding. Then we demonstrate the results of the entire navigation system in both simulated and real-world scenarios. 

\subsection{Phrase Classification}
The phrase classification accuracy is very important for the following grounding operations. Thus in our experiment, we have tested different networks and compared their performance. In particular, we test three different networks: the standard LSTM network (LSTM), the bi-directional LSTM network (Bi-LSTM), and the bi-directional LSTM with attention mechanism (Att-Bi-LSTM). All these networks are trained using the Adam optimizer with the cross-entropy as the loss function. 

\begin{figure}[t] 
\centering
\includegraphics[trim=32 0 45 10, clip, width=\linewidth]{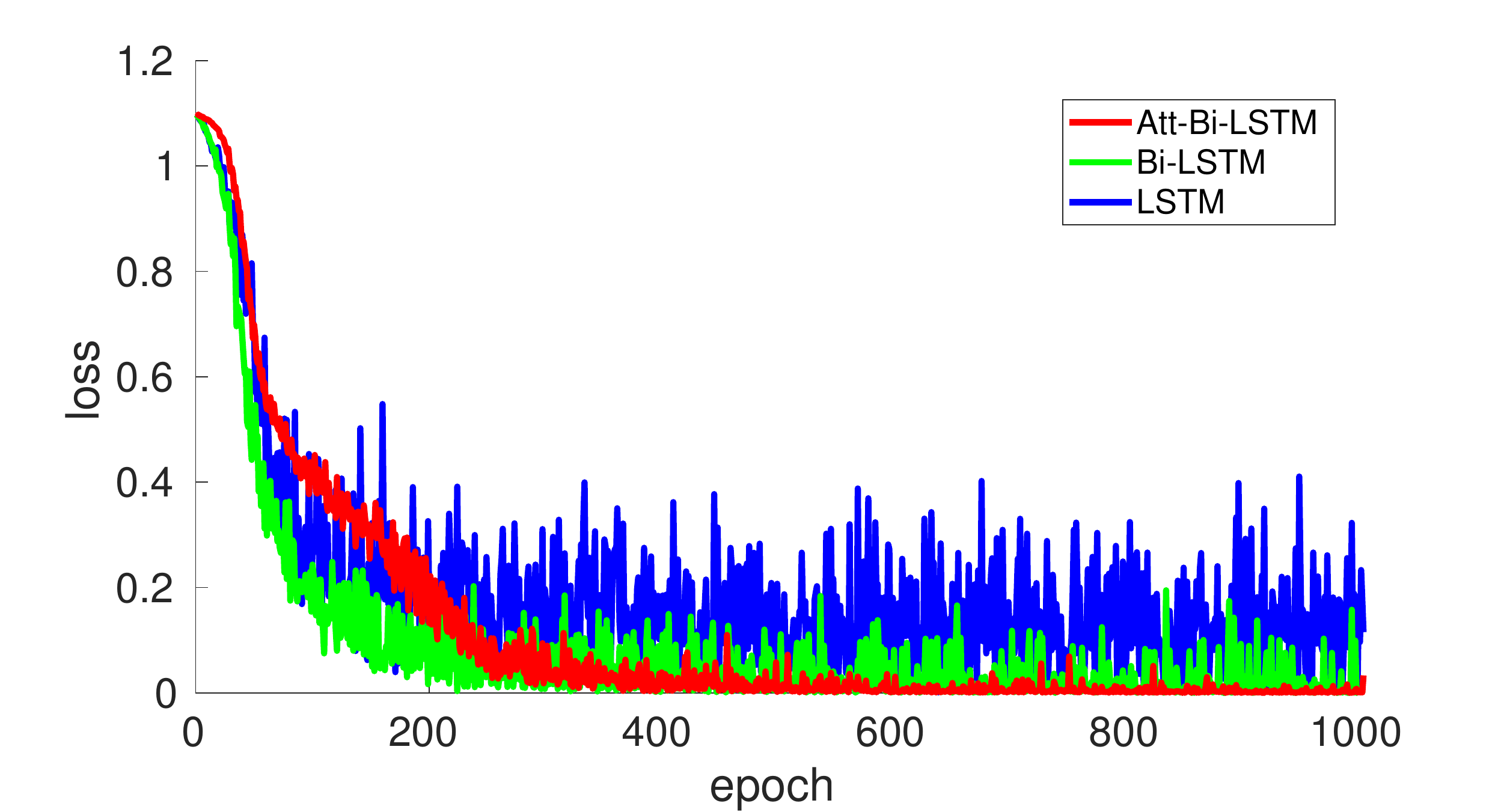}
\caption{The loss values in epochs during the training process for three LSTM networks: the standard LSTM (LSTM), the bi-directional LSTM (Bi-LSTM), and the bi-directional LSTM with attention mechanism (Att-Bi-LSTM).}
\label{fig:loss}
\end{figure}

The comparison of the training losses of the three LSTM networks is shown in \prettyref{fig:loss}. We can observe that all three LSTM networks are able to reach a relatively low loss value, which implies that they can learn how to classify the instruction phrases correctly. The bi-directional LSTM with attention mechanism provides the best performance with a loss value of around 0.006, which is much lower than that of the standard LSTM (around 0.15) and the bi-directional LSTM (around 0.05).

To perform training and testing for phrase classification, we collect a dataset containing 500 navigation instructions. The accuracy of the three LSTM networks mentioned above is tested in a test set with 300 phrases, and the result is shown in \prettyref{tab:accuracy}. We can observe that the bi-directional LSTM with attention mechanism achieves the best performance with 96\% accuracy. As a result, in the following experiments, we choose the bi-directional LSTM with attention mechanism as our phrase classification network.


\begin{table}[t]
\resizebox{0.48\textwidth}{!}{
\begin{tabular}
{c|c|c|c}
    \hline
     Network & LSTM  & Bi-LSTM & Att-Bi-LSTM \\ \hline
     Accuracy & 82\% & 88\% & 96\%\\ \hline
\end{tabular}
 }
\caption{The accuracy for phrase classification of three LSTM architectures, including the standard LSTM (LSTM), the bi-directional LSTM (Bi-LSTM), and the bi-directional LSTM with attention mechanism (Att-Bi-LSTM).}
\label{tab:accuracy}
\vspace*{-0.2in}
\end{table}

\subsection{Dynamic Constraint Grounding}
To recognize the constraint objects mentioned in the human instructions and ground their locations, we use the mask R-CNN network~\cite{he2017mask} to perform instance segmentation. The network is trained on a COCO dataset which includes 80 different objects or labels. \prettyref{fig:seg} shows the output of the mask R-CNN. To compute the location of the detected constraint objects, we use the mask produced by the mask R-CNN to crop the depth image from the RGB-D sensor, and then compute the average depth value of each masked depth image. The average depth value is then converted from the camera's coordinate to the robot's local coordinate to provide the location of the constraint objects. After that, the similarity between the constraint objects mentioned in the instruction and the labels provided by the instance segmentation is computed to accomplish the grounding process. 

\begin{figure}[t] 
\centering
\includegraphics[trim=40 0 45 10, clip, width=\linewidth]{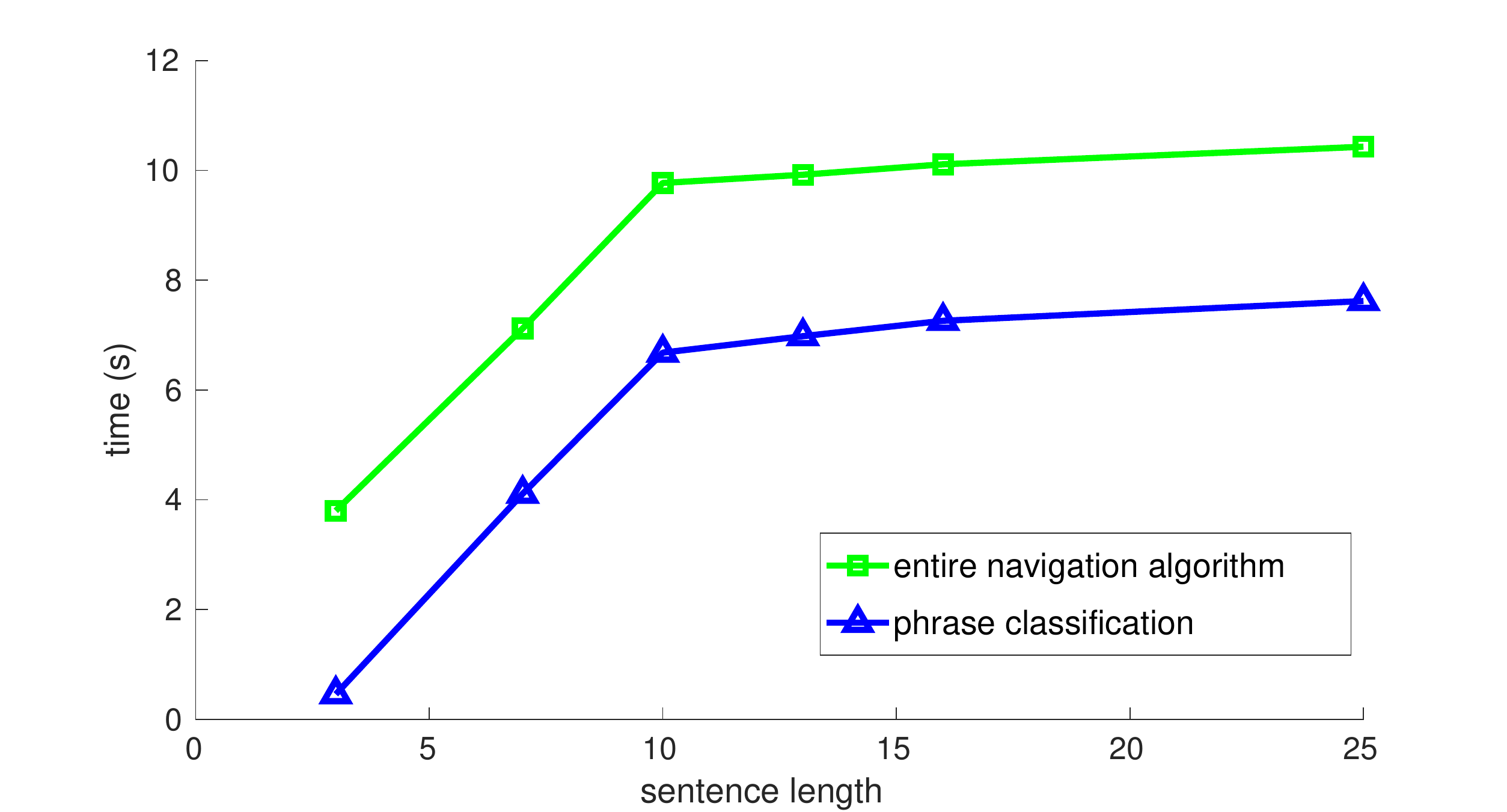}
\caption{Both the phrase classification and the entire navigation pipeline require more computational time when the human instruction becomes longer in sentences. }
\label{fig:time}
\end{figure}

\begin{figure}[t] 
\centering
\begin{subfigure}{\linewidth}
\includegraphics[trim=0 60 0 60, clip, width=\linewidth]{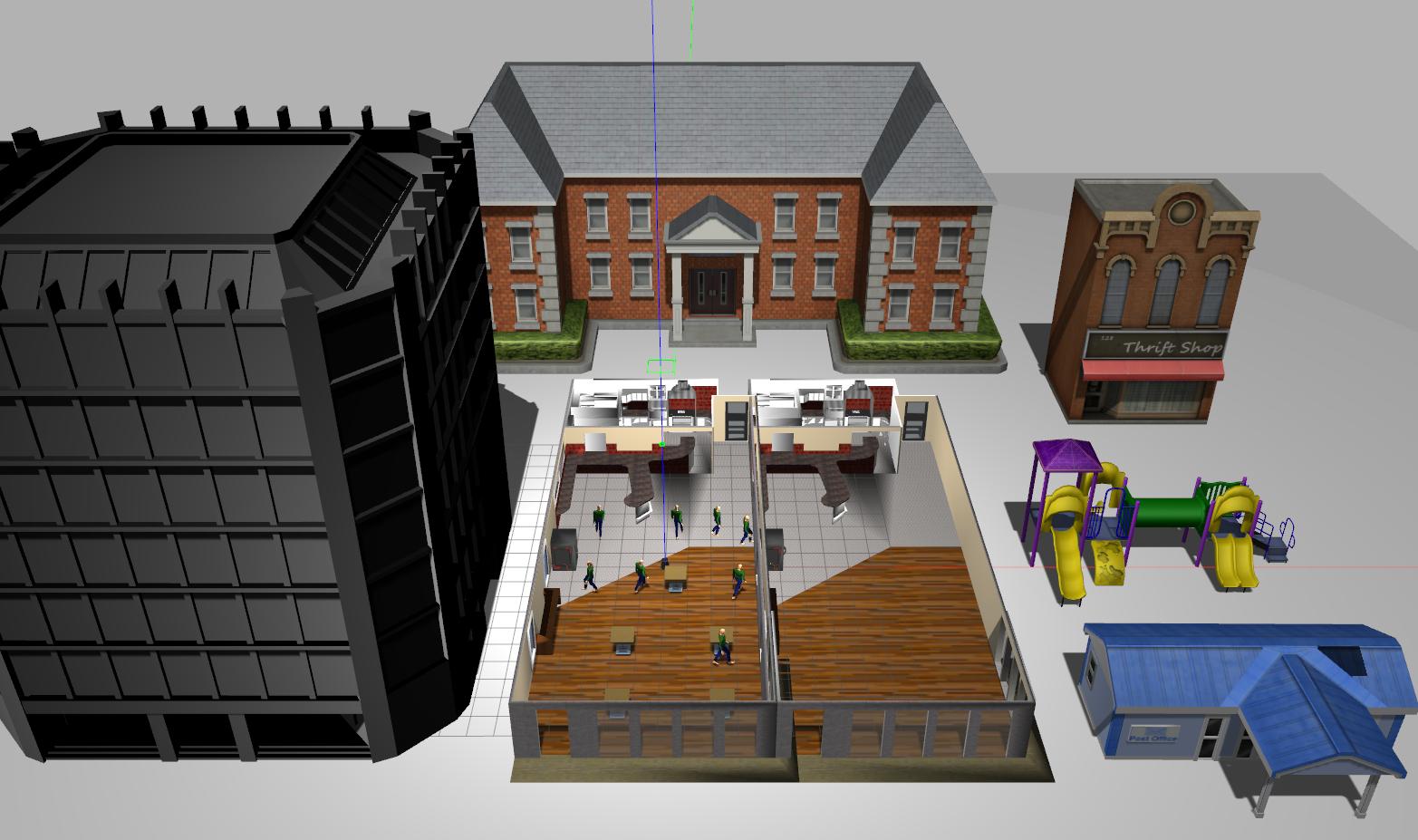}
\caption{}
\end{subfigure}
\begin{subfigure}{0.49\linewidth}
\includegraphics[trim=0 50 0 10, clip, width=1.0\linewidth]{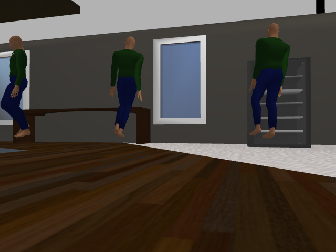}
\caption{}
\label{fig:rgb}
\end{subfigure}
\begin{subfigure}{0.49\linewidth} 
\centering
\includegraphics[trim=0 50 0 10, clip, width=1.0\linewidth]{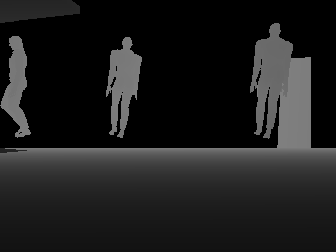}
\caption{}
\label{fig:depth}
\end{subfigure}
\caption{We test our navigation algorithm in a simulated environment with static obstacles and moving pedestrians, as shown (a). The simulated robot can obtain the RGB and depth information about the scene, as shown in (b) and (c) respectively. Please refer to the video for more details about the simulation.}
\label{fig:sim}
\vspace*{-0.2in}
\end{figure}

\begin{table}[t]
\small
\resizebox{0.48\textwidth}{!}{
{\renewcommand{\arraystretch}{2}
\begin{tabular}
{p{7cm}|l|l }
    \hline
     \bf instructions & \bf goal  & \bf constraint object  \\ \hline \hline
     ``go to the restaurant and you know, keep away from people.'' & ``restaurant'' & ``people'' \\ \hline
     ``move to the laboratory and watch out the table and chairs.'' & laboratory & ``table'' ``chair''  \\ \hline
     ``robot, go to the lift'' & ``lift'' &   \\ \hline
     ``don't collide with people and walk to the information desk'' & ``information desk'' & ``people''   \\ \hline 
    ``robot, go to the school and stay away from children'' & ``school'' & ``children''    \\ \hline
    ``go to the thrift shop to buy some water and watch out the table in the shop and you know, keep away from the people'' & ``thrift shop'' & ``table'' ``people'' \\ \hline
  \end{tabular}
 }
 }
\caption{Samples of natural language human instructions used in our simulated environment. The second and third columns are the goal and constraint objects extracted using our phrase classification and grounding algorithm.}
\label{tab:ins}
\end{table}

\begin{table}[t]
\resizebox{0.48\textwidth}{!}{
\begin{tabular}
{P{1.2cm}|c|c|c|c|c|c|c}
    \hline
     \#constraints & 0 & 1 & 2 & 3 & 4 & 5 & 6\\ \hline
     path length & 2.83 & 3.02 &3.13&3.42&3.42&3.45&3.52 \\ \hline
     time & 5.3 & 6.4 &6.8&7.1&7.3&7.4&7.6 \\ \hline
  \end{tabular}
 }
\caption{We compare the length (in meters) and execution time (in seconds) of the paths generated by our navigation system given the different number of constraints and the same goal.}
\label{tab:path_length}
\vspace*{-0.2in}
\end{table}

\subsection{Navigation Performance}
We test our algorithm using a set of navigation instructions shown in \prettyref{tab:ins}. We also record the execution time of the entire system with respect to the different length of the instruction sentences, and the result is shown in \prettyref{fig:time}.

\subsubsection{Simulated Environments}
As shown in \prettyref{fig:sim}, we build a complex simulated scene to test the performance of our algorithm. This scenario covers both the indoor and outdoor environments. It contains static obstacles such as ``office building'', ``post office'', ``cafe'', ``thrift shop'', ``school'', ``table'' and ``playground''. It also contains moving obstacles like walking pedestrians in the ``cafe'' region. 

Our navigation algorithm can achieve successful navigation in such complex scenarios by following human instructions. Please refer to the video for the details. In addition, we also evaluate how the path planning performance changes given the different number of constraints. We fix the goal position and gradually add more constraints. The lengths and the execution time of the resulting trajectories are listed in \prettyref{tab:path_length}. We can see that both the trajectory length and execution time increase when the number of constraints increases, but the first two constraints will bring the most significant changes.

\subsubsection{Real-world Environment}
\begin{figure}[t] 
\centering
\begin{subfigure}{\linewidth}
\includegraphics[trim=15 55 30 20, clip,width=1.0\linewidth]{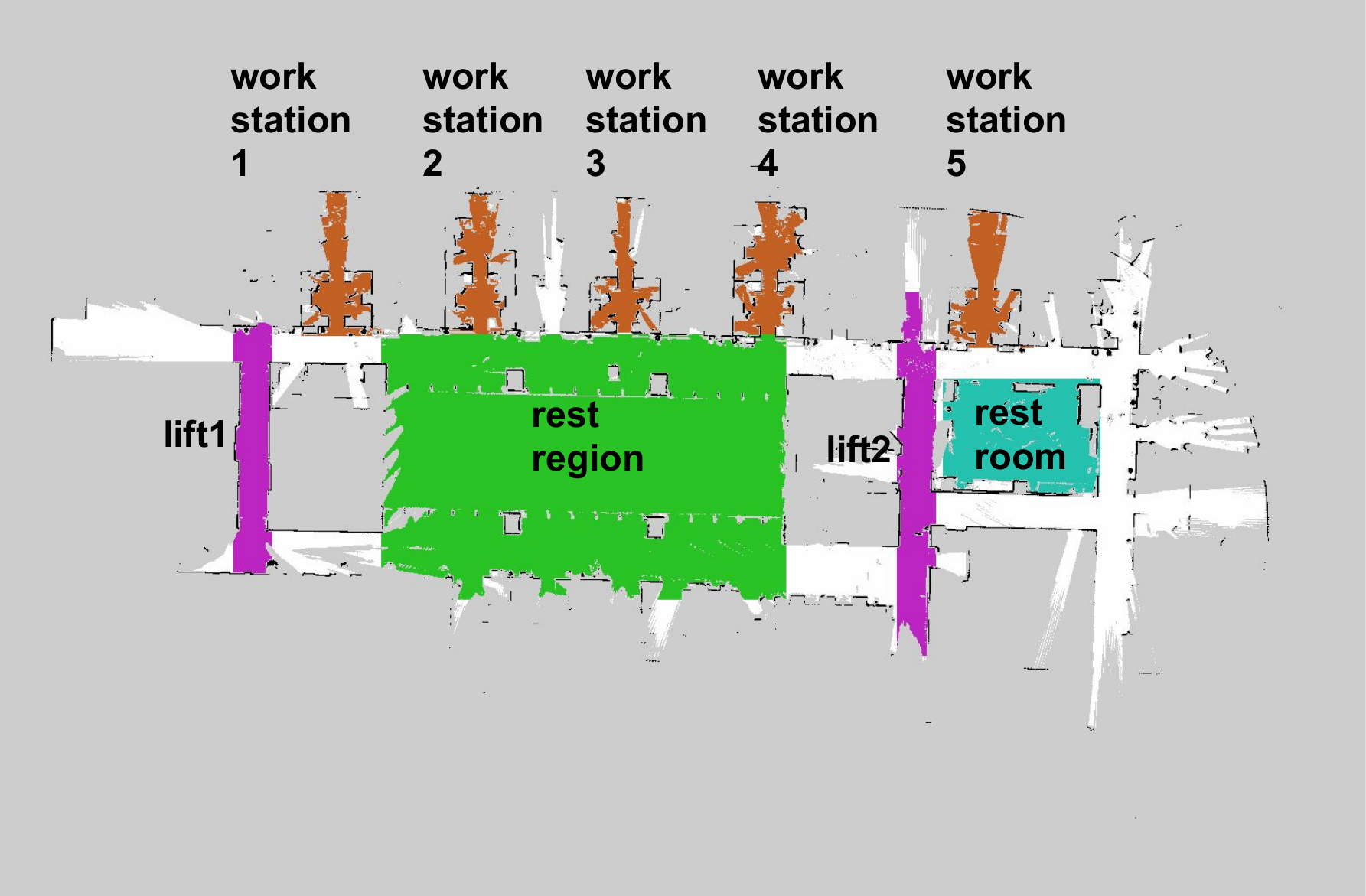}
\caption{semantic map for scene 1}
\label{fig:map1}
\end{subfigure}
\begin{subfigure}{\linewidth}
\centering
\includegraphics[trim=15 35 30 65, clip,width=1.0\linewidth]{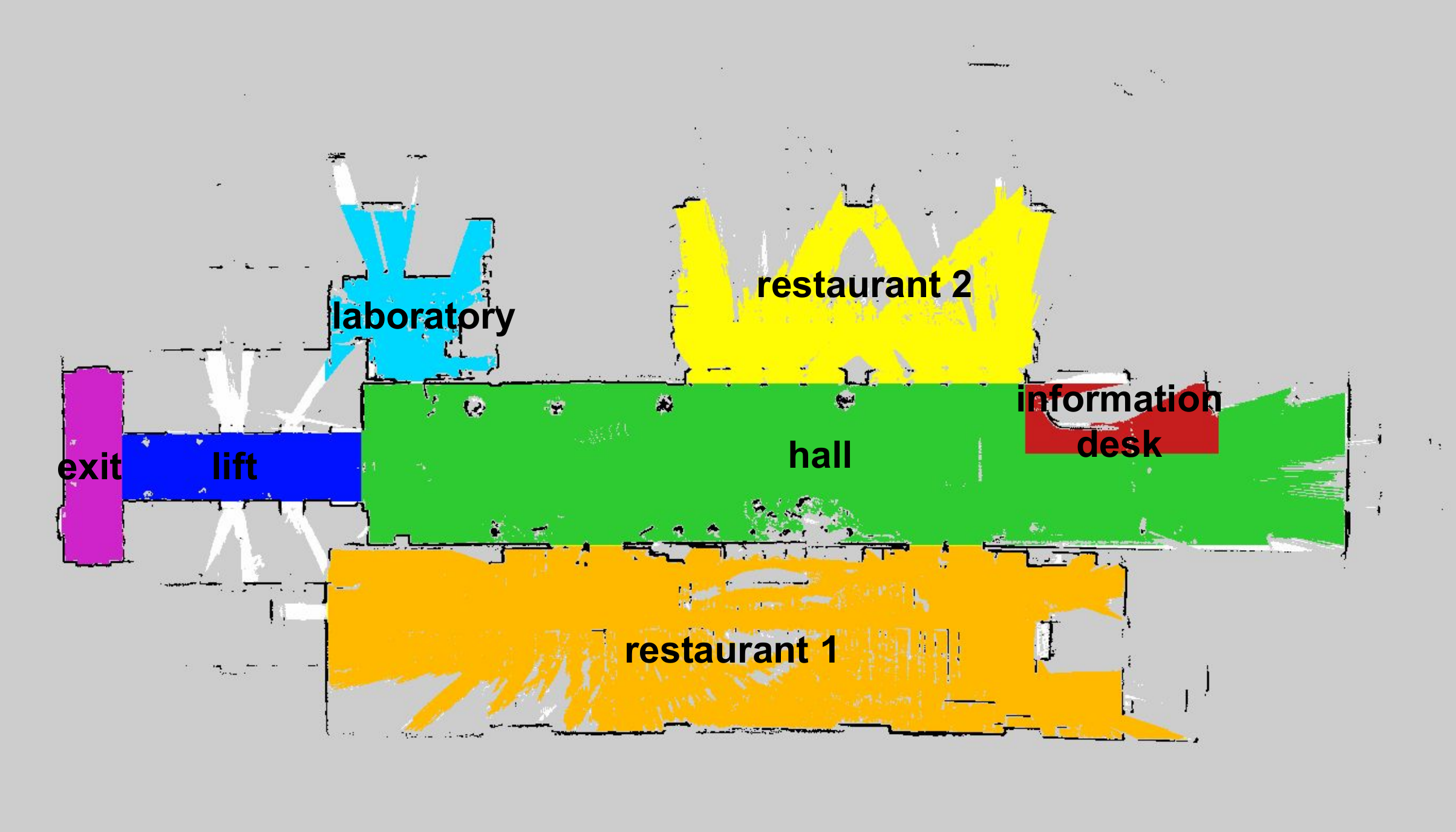}
\caption{semantic map for scene 2}
\label{fig:map2}
\end{subfigure}
\caption{Two semantic maps for different scenarios constructed using SLAM.}
\label{fig:map}
\vspace*{-0.2in}
\end{figure}

For the real-world experiment, we first run the SLAM algorithm~\cite{cartographer} around our lab to construct a global amp, and then manually annotate this map with semantic information. The resulting semantic maps are shown in \prettyref{fig:map}, including semantic locations such as restaurants, information desks, laboratory, lifts, hall, rest regions, and workstations. Our navigation algorithm can enable a robot to follow human's navigation instructions successfully in these scenarios. Please refer to the video for more details. 

\begin{figure}[t] 
\centering
\begin{subfigure}{\linewidth}
\includegraphics[trim=15 60 30 35, clip,width=1.0\linewidth]{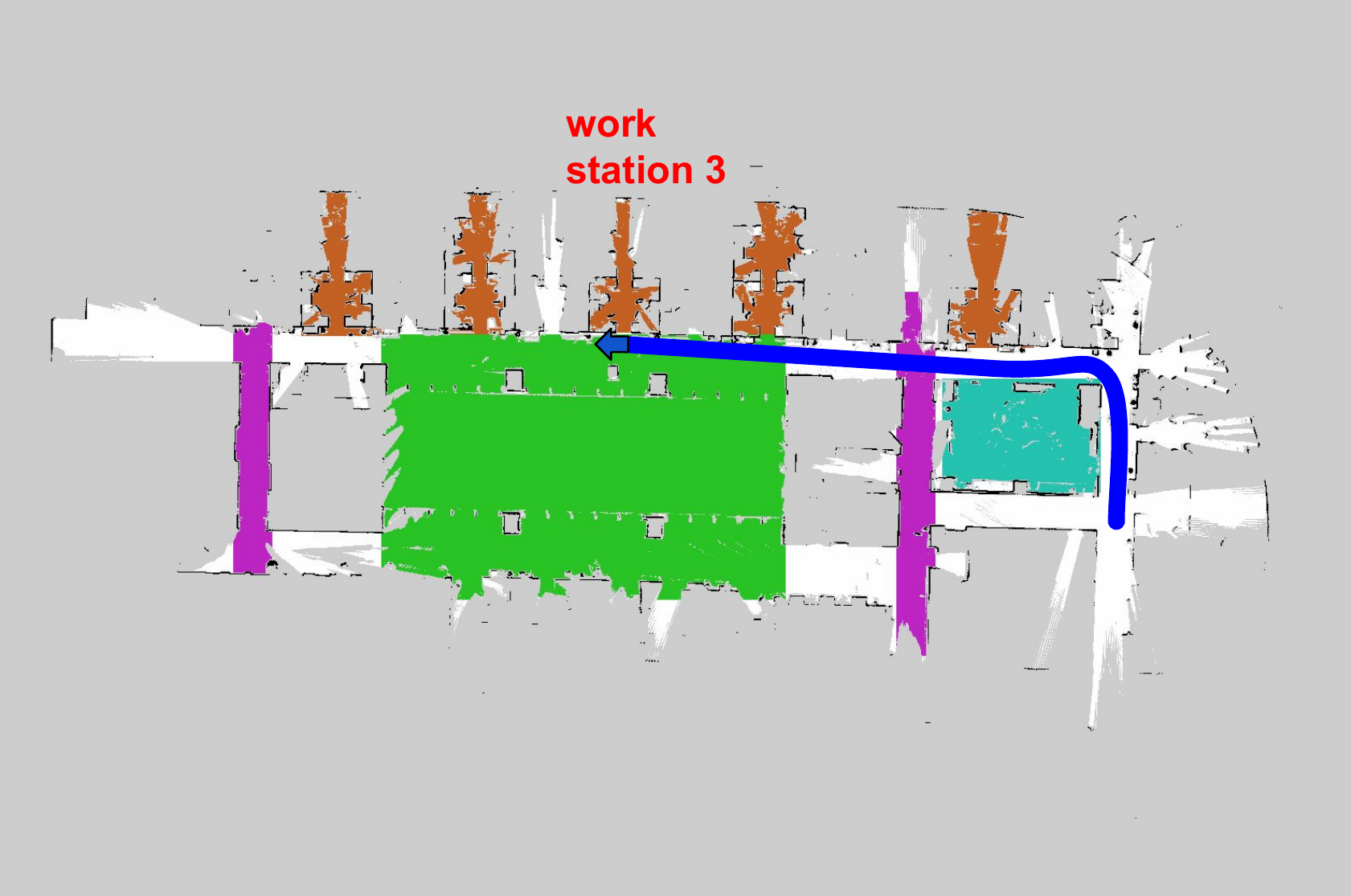}
\caption{path planning without constraints}
\label{fig:planning1}
\end{subfigure}
\begin{subfigure}{\linewidth} 
\centering
\includegraphics[trim=15 60 30 35, clip,width=1.0\linewidth]{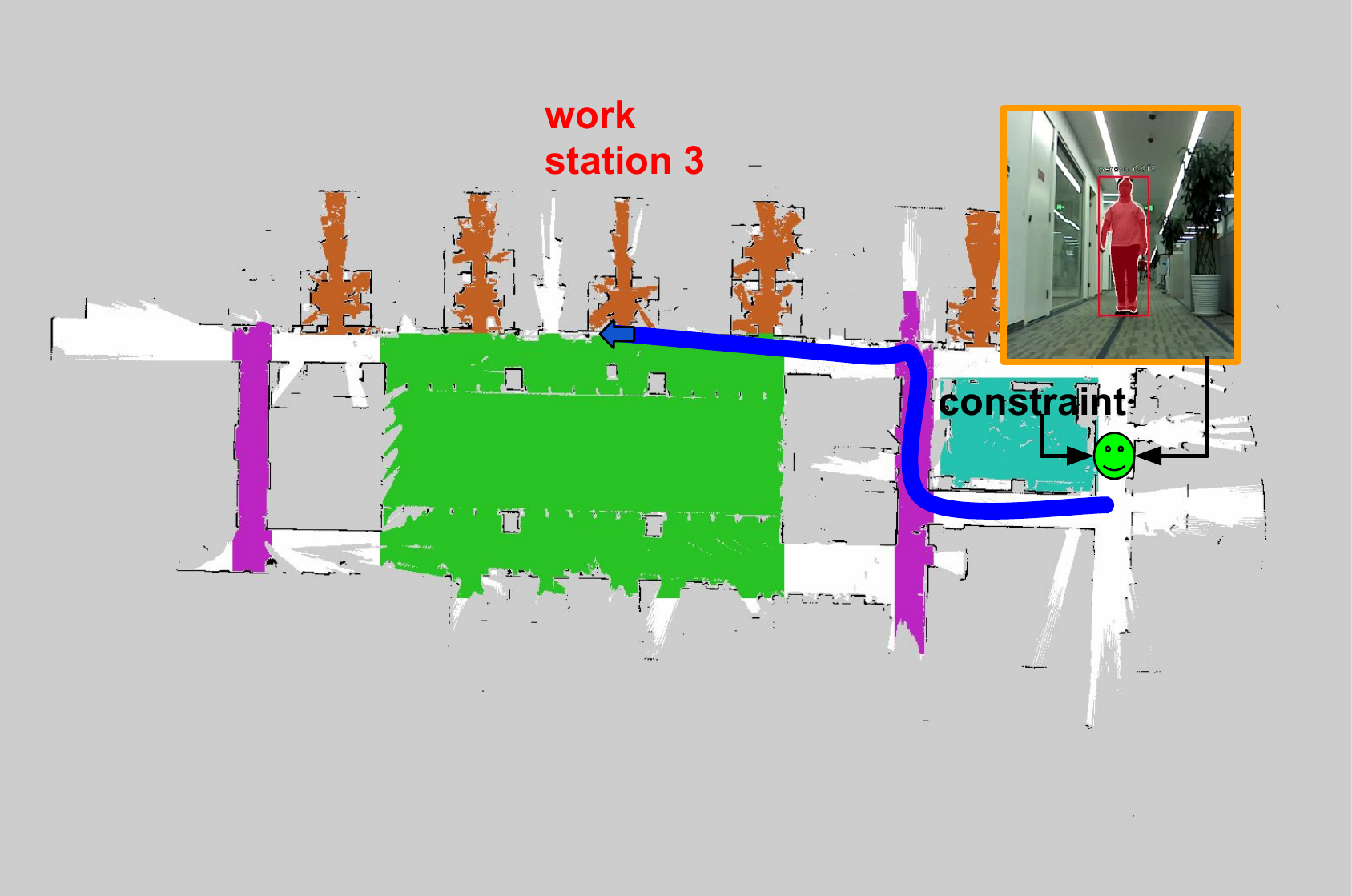}
\caption{path planning with the constraint object ``person''}
\label{fig:planning2}
\end{subfigure}
\caption{Comparison of the planning results without and with constraints. In (a), the human constraint object is not taken into account in the planning, and thus it will choose a path close to the humans. In (b), the planner will recognize the human constraint object and take an alternative path to avoid humans.}
\label{fig:planning}
\vspace*{-0.2in}
\end{figure}

To demonstrate the impact of constraint grounding, in the scenario shown in \prettyref{fig:map1}, we run a contrast experiment comparing the navigation trajectory without and with constraints. \prettyref{fig:planning1} shows the navigation result when receiving the instruction ``go to the restaurant,''  and \prettyref{fig:planning2} shows the trajectory when receiving ``go to the restaurant and you know, keep away from people''. We can observe that the existence of constraints significantly changes the robot's trajectory. 

\section{Conclusion and Future work}
\label{sec:conclusion}
In this paper, we solve the problem of robotic navigation following a complex natural language instruction by combining natural language processing, computer vision, local/global motion planning, and collision avoidance techniques. By using the costmap motion planner as a bridge between the natural language instruction and the collision-free navigation command, our system accomplishes safe navigation in complex scenarios.  

For future work, we plan to train an end-to-end neural network to combine all the components proposed in this work, including phrase classification, dynamic constraint grounding, and obstacle avoidance, to allow all the parameters to be optimized by the robot's navigation experience.

{\small
\bibliographystyle{IEEEtran}
\bibliography{references}

\begin{thebibliography}{10}
\providecommand{\url}[1]{#1}
\csname url@samestyle\endcsname
\providecommand{\newblock}{\relax}
\providecommand{\bibinfo}[2]{#2}
\providecommand{\BIBentrySTDinterwordspacing}{\spaceskip=0pt\relax}
\providecommand{\BIBentryALTinterwordstretchfactor}{4}
\providecommand{\BIBentryALTinterwordspacing}{\spaceskip=\fontdimen2\font plus
\BIBentryALTinterwordstretchfactor\fontdimen3\font minus
  \fontdimen4\font\relax}
\providecommand{\BIBforeignlanguage}[2]{{%
\expandafter\ifx\csname l@#1\endcsname\relax
\typeout{** WARNING: IEEEtran.bst: No hyphenation pattern has been}%
\typeout{** loaded for the language `#1'. Using the pattern for}%
\typeout{** the default language instead.}%
\else
\language=\csname l@#1\endcsname
\fi
#2}}
\providecommand{\BIBdecl}{\relax}
\BIBdecl

\bibitem{wei2009go}
Y.~Wei, E.~Brunskill, T.~Kollar, and N.~Roy, ``Where to go: Interpreting
  natural directions using global inference,'' in \emph{ICRA}, 2009, pp.
  3761--3767.

\bibitem{kollar2014grounding}
T.~Kollar, S.~Tellex, D.~Roy, and N.~Roy, ``Grounding verbs of motion in
  natural language commands to robots,'' in \emph{Experimental robotics}, 2014,
  pp. 31--47.

\bibitem{kollar2010toward}
------, ``Toward understanding natural language directions,'' in \emph{HRI},
  2010, pp. 259--266.

\bibitem{posada2014visual}
L.~F. Posada, F.~Hoffmann, and T.~Bertram, ``Visual semantic robot navigation
  in indoor environments,'' in \emph{ISR}, 2014, pp. 1--7.

\bibitem{kollar2013generalized}
T.~Kollar, S.~Tellex, M.~R. Walter, A.~Huang, A.~Bachrach, S.~Hemachandra,
  E.~Brunskill, A.~Banerjee, D.~Roy, S.~Teller \emph{et~al.}, ``Generalized
  grounding graphs: A probabilistic framework for understanding grounded
  language,'' \emph{JAIR}, 2013.

\bibitem{howard2014natural}
T.~M. Howard, S.~Tellex, and N.~Roy, ``A natural language planner interface for
  mobile manipulators,'' in \emph{ICRA}, 2014, pp. 6652--6659.

\bibitem{arkin2015towards}
J.~Arkin and T.~M. Howard, ``Towards learning efficient models for natural
  language understanding of quantifiable spatial relationships,'' in \emph{RSS
  2015 Workshop on Model Learning for Human-Robot Communication}, 2015.

\bibitem{chung2015performance}
I.~Chung, O.~Propp, M.~R. Walter, and T.~M. Howard, ``On the performance of
  hierarchical distributed correspondence graphs for efficient symbol grounding
  of robot instructions,'' in \emph{IROS}, 2015, pp. 5247--5252.

\bibitem{paul2016efficient}
R.~Paul, J.~Arkin, N.~Roy, and T.~M~Howard, ``Efficient grounding of abstract
  spatial concepts for natural language interaction with robot manipulators,''
  in \emph{RSS}, 2016.

\bibitem{park2017generating}
J.~S. Park, B.~Jia, M.~Bansal, and D.~Manocha, ``Generating realtime motion
  plans from complex natural language commands using dynamic grounding
  graphs,'' \emph{arXiv:1707.02387}, 2017.

\bibitem{huang2018finding}
D.-A. Huang, S.~Buch, L.~Dery, A.~Garg, L.~Fei-Fei, and J.~C. Niebles,
  ``Finding “it”: Weakly-supervised reference-aware visual grounding in
  instructional videos,'' in \emph{CVPR}, 2018.

\bibitem{shridhar2018interactive}
M.~Shridhar and D.~Hsu, ``Interactive visual grounding of referring expressions
  for human-robot interaction,'' \emph{arXiv:1806.03831}, 2018.

\bibitem{tellex2011understanding}
S.~Tellex, T.~Kollar, S.~Dickerson, M.~R. Walter, A.~G. Banerjee, S.~J. Teller,
  and N.~Roy, ``Understanding natural language commands for robotic navigation
  and mobile manipulation,'' in \emph{AAAI}, vol.~1, 2011, p.~2.

\bibitem{matuszek2010following}
C.~Matuszek, D.~Fox, and K.~Koscher, ``Following directions using statistical
  machine translation,'' in \emph{HRI}, 2010, pp. 251--258.

\bibitem{duvallet2013imitation}
F.~Duvallet, T.~Kollar, and A.~Stentz, ``Imitation learning for natural
  language direction following through unknown environments,'' in \emph{ICRA},
  2013, pp. 1047--1053.

\bibitem{anderson2018vision}
P.~Anderson, Q.~Wu, D.~Teney, J.~Bruce, M.~Johnson, N.~S{\"u}nderhauf, I.~Reid,
  S.~Gould, and A.~van~den Hengel, ``Vision-and-language navigation:
  Interpreting visually-grounded navigation instructions in real
  environments,'' in \emph{CVPR}, 2018.

\bibitem{wu2018building}
Y.~Wu, Y.~Wu, G.~Gkioxari, and Y.~Tian, ``Building generalizable agents with a
  realistic and rich 3d environment,'' \emph{arXiv:1801.02209}, 2018.

\bibitem{matuszek2012joint}
C.~Matuszek, N.~FitzGerald, L.~Zettlemoyer, L.~Bo, and D.~Fox, ``A joint model
  of language and perception for grounded attribute learning,''
  \emph{arXiv:1206.6423}, 2012.

\bibitem{hochreiter1997long}
S.~Hochreiter and J.~Schmidhuber, ``Long short-term memory,'' \emph{Neural
  computation}, vol.~9, no.~8, pp. 1735--1780, 1997.

\bibitem{long2017towards}
P.~Long, T.~Fan, X.~Liao, W.~Liu, H.~Zhang, and J.~Pan, ``Towards optimally
  decentralized multi-robot collision avoidance via deep reinforcement
  learning,'' in \emph{ICRA}, 2018.

\bibitem{graves2005framewise}
A.~Graves and J.~Schmidhuber, ``Framewise phoneme classification with
  bidirectional lstm and other neural network architectures,'' \emph{Neural
  Networks}, vol.~18, no. 5-6, pp. 602--610, 2005.

\bibitem{mikolov2013efficient}
T.~Mikolov, K.~Chen, G.~Corrado, and J.~Dean, ``Efficient estimation of word
  representations in vector space,'' \emph{arXiv:1301.3781}, 2013.

\bibitem{mikolov2013distributed}
T.~Mikolov, I.~Sutskever, K.~Chen, G.~S. Corrado, and J.~Dean, ``Distributed
  representations of words and phrases and their compositionality,'' in
  \emph{NIPS}, 2013, pp. 3111--3119.

\bibitem{cartographer}
W.~Hess, D.~Kohler, H.~Rapp, and D.~Andor, ``Real-time loop closure in 2d lidar
  slam,'' in \emph{ICRA}, 2016, pp. 1271--1278.

\bibitem{schulman2017proximal}
J.~Schulman, F.~Wolski, P.~Dhariwal, A.~Radford, and O.~Klimov, ``Proximal
  policy optimization algorithms,'' \emph{arXiv:1707.06347}, 2017.

\bibitem{he2017mask}
K.~He, G.~Gkioxari, P.~Doll{\'a}r, and R.~Girshick, ``Mask {R-CNN},'' in
  \emph{ICCV}, 2017, pp. 2980--2988.

\end{thebibliography}
}

\end{document}